\documentclass[journal]{IEEEtran}

\usepackage{graphicx}
\usepackage{caption}
\usepackage{subcaption}
\usepackage{algorithm}
\usepackage{algpseudocode}
\usepackage{amsmath,amsfonts}
\usepackage{multirow}

\ifCLASSINFOpdf
\else
\fi

\begin{document}

\title{Large-scale Mixed Traffic Control Using Dynamic Vehicle Routing and Privacy-Preserving Crowdsourcing}

\author{Dawei Wang, Weizi Li, Jia Pan 
\thanks{Dawei Wang and Jia Pan are with both the Department of Computer Science and TransGP, The University of Hong Kong. Weizi Li is with the Min H. Kao Department of Electrical Engineering and Computer Science, University of Tennessee, Knoxville. Jia Pan is the corresponding author.
        {\tt\footnotesize jpan@cs.hku.hk}}
}

\markboth{Journal of \LaTeX\ Class Files}%
{Shell \MakeLowercase{\textit{et al.}}: Large-scale Mixed Traffic Control Using Dynamic Vehicle Routing and Privacy-Preserving Crowdsourcing }

\maketitle

\begin{abstract}
Controlling and coordinating urban traffic flow through robot vehicles is emerging as a novel transportation paradigm for the future.
While this approach garners growing attention from researchers and practitioners, effectively managing and coordinating large-scale mixed traffic remains a challenge.
We introduce an effective framework for large-scale mixed traffic control via  privacy-preserving crowdsourcing and dynamic vehicle routing. 
Our framework consists of three modules: a privacy-protecting crowdsensing method, a graph propagation-based traffic forecasting method, and a privacy-preserving route selection mechanism. We evaluate our framework using a real-world road network.
The results show that our framework accurately forecasts traffic flow, efficiently mitigates network-wide RV shortage issue, and coordinates large-scale mixed traffic. 
Compared to other baseline methods, our framework not only reduces the RV shortage issue up to $69.4\%$ but also reduces the average waiting time of all vehicles in the network up to $27\%$. 
\end{abstract}

\begin{IEEEkeywords}
Mixed Traffic Control, Multi-agent Reinforcement Learning, Privacy, Crowdsourcing, Future Mobility 
\end{IEEEkeywords}

\IEEEpeerreviewmaketitle

\section{Introduction}
\IEEEPARstart{W}{ith} the rapid advancement of autonomous driving technology, an escalating number of vehicles are now being equipped with autonomy capabilities and tested in real-world traffic.
Anticipating the decades ahead, the gradual shift towards full transportation autonomy will usher in an era of \emph{mixed traffic}, where both human-driven vehicles (HVs) and robot vehicles (RVs) coexist, equipped with diverse levels of autonomous driving features. 
Consequently, an emerging concern for our transportation systems is the effective management and coordination of mixed traffic.

Despite the challenges posed by the diverse and suboptimal nature of human drivers in mixed traffic, recent research has demonstrated that reinforcement learning (RL) can serve as a promising approach to acquire adaptive behaviors for RVs, with the primary goal of optimizing overall traffic efficiency by influencing nearby HVs~\cite{wu2021flow,yan2021reinforcement,wang2023learning,Villarreal2023Pixel,niroumand2023white}.
In particular, researchers have shown that employing RVs as the sole control mechanism in mixed traffic can outperform traditional traffic lights at unsignalized intersections~\cite{yan2021reinforcement,wang2023learning}.
However, existing methods are predominantly tailored for simple road networks (including single intersections) and are unsuitable for managing large-scale mixed traffic. 
To be specific, these traffic control methods necessitate a minimum RV presence to realize the advantages of mixed traffic in enhancing overall traffic efficiency.
Consequently, one challenge in large-scale mixed traffic control within complex road networks is the need to maintain a delicate equilibrium in RV penetration rates, which fluctuate due to traffic dynamics and unpredictable behaviors of HVs. 
Failing to maintain this balance may lead to congestion or even network-wide gridlock. 
As of now, \emph{there lacks of an RV routing algorithm to guarantee such balanced RV rates across road segments of a road network}.

Another challenge that hinders the control of large-scale mixed traffic is the perception of individual robot vehicles through on-board sensors such as GPS, radars, and LiDARs. 
They are limited in obtaining a global view of the road network and the associated traffic conditions.
One approach to mitigate this issue is to utilize Internet of Vehicles (IoV). 
IoV constitutes an interactive network where vehicles can exchange information with each other and with the infrastructure. 
This development holds the promise of enabling RVs to leverage not only local perception but also global traffic information about a road network. 
In our context, one way to utilize IoV is to allow a group of RVs form a crowdsourcing network to jointly collect traffic information by using a central server to route the RVs according to their destinations and network traffic conditions. 
While this approach is straightforward, privacy concerns arise because information, such as destinations and planned routes, must be shared with the central server, along with other data collected by neighboring RVs about HVs. 
This raises two crucial aspects of privacy preservation in large-scale traffic coordination: perception—--\emph{how to collect traffic data while preserving the privacy of other vehicles}, and routing—--\emph{how to route RVs while preserving their private information, such as trajectories}.

While some methods~\cite{ni2016privacy,kumar2021privacy} have been proposed to protect user privacy in the context of IoV and autonomous driving, to the best of our knowledge, there has been no study showing the practicality of managing and coordinating mixed traffic across extensive real-world road networks. This stands as a pivotal evaluation of whether autonomous driving can make a positive impact on urban environments. 
In this project, we propose an effective method for large-scale mixed traffic control. 
Our method involves a novel vehicle routing algorithm that incorporates a privacy-preserving mechanism for gathering network-wide traffic information. 
Our contributions include: 
\begin{itemize}
    \item A novel framework for large-scale mixed traffic control and coordination; 
    \item A privacy-protecting crowdsensing method for collecting traffic conditions via RVs;
    \item A graph propagation-based method for modeling and predicting traffic flow;
     and
    \item A privacy-preserving crowdsourcing mechanism for route selection of RVs while balancing RV penetration rates across road segments of a road network.  
\end{itemize} 

We evaluate the proposed framework using simulations constructed in the city of Colorado Springs, CO, USA. 
The results show that our method can effectively control and coordinate large-scale mixed traffic on road networks even without the presence of traffic lights, and achieve better performance compared to baseline methods as the RV penetration rate increases.  
For example, our method can reduce the average RV shortage index by 69.4\% compared to a baseline method~\cite{wang2023learning}. 
As a result of alleviating the RV shortage situation, traffic efficiency is significantly improved, for example, compared to Wang et al.~\cite{wang2023learning}, our method can reduce the average waiting time of vehicles by 27\% under 50\% RV penetration rate.

\section{Related Work}
We provide a brief overview of related studies pertaining to various components of our framework.

\subsection{Traffic State Prediction and Estimation} 
\label{sec:review-forecasting}

Li et al.~\cite{li2017diffusion} categorize traffic state prediction methods into two groups: knowledge-driven approaches and data-driven approaches. As an example of knowledge-driven approach, Cascetta et al.~\cite{cascetta2013transportation} propose a method employing queuing theory and behavior simulation. In recent years, data-driven methods have received considerable attention. For example, Liu et al.~\cite{liu2011discovering} introduce an Auto-Regressive Integrated Moving Average (ARIMA) model for modeling and predicting traffic data. Lippi et al.~\cite{lippi2013short} propose Seasonal Auto-Regressive Integrated Moving Average (SARIMA) model coupled with Kalman filter. 
Ma et al.~\cite{ma2017learning} introduce a CNN-based method that converts traffic data to images and predict large-scale traffic states. Li et al.~\cite{li2017diffusion} introduce a diffusion convolutional recurrent network that leverages directed graphs for traffic flow forecasting. 
Lin et al.~\cite{Lin2019Compress,Lin2019BikeTRB,Lin2022GCGRNN} propose a series of traffic state prediction algorithms that are data-efficient, and multi-step and network-wide. 
Li et al.~\cite{Li2017CitySparseITSM, Li2018CityEstIET} develop methods to estimate city-scale traffic states using sparse GPS data.  
While significant progress has been made, existing methods are not designed for mixed traffic, especially in predicting short-term RV penetration rate across the road network.

\subsection{Vehicle Route Planning}
\label{sec:review-planning}
Ge et al.~\cite{ge2010energy} present a system for recommending routes that uses a travel distance function to extract energy-efficient patterns and evaluate candidate sequences. Shang et al.~\cite{shang2015collective} propose a novel method for planning collective travel that finds the lowest cost route connecting multiple query sources. Li et al.~\cite{li2005trip} introduce a query method for trip planning to determine the best possible route. Zhang et al.~\cite{zhang2022route} develope a framework for route planning that considers multiple objectives such as travel distance, travel time, cost, and traveler preferences. 
Since mixed traffic control is a new development, a method is required to consider RV penetration rates, as they can significantly affect traffic efficiency in mixed traffic control~\cite{wang2023learning}.

\subsection{Mixed Traffic Control at Intersections}
\label{sec:review-traffic-control}
Traffic signals are the primary source of controlling traffic at intersections. However, RVs have the potential to control traffic without the need for traffic signals. Sharon et al.~\cite{sharon2017protocol} introduce the Hybrid Autonomous Intersection Management protocol, which uses sensing data from infrastructure to manage traffic. Yang et al.~\cite{yang2020intelligent} predict vehicle delay and find the optimal control action for connected vehicles, improving the performance of the traffic system. Wu
 et al. use reinforcement learning~\cite{vinitsky2018lagrangian,wu2021flow} for mixed traffic control. Yan et al.~\cite{yan2021reinforcement} also leverages reinforcement learning to control mixed traffic at intersections. Recently, Wang et al.~\cite{wang2023learning} leverages reinforcement learning to control and coordinate mixed traffic at complex and unsignalized intersections, which we adopt as the intersectional traffic control mechanism for this project. 

\subsection{Privacy Preservation}
The development of Internet of Vehicles (IoV) has led to advancements in intelligent transportation systems, improving road safety, traffic efficiency, and driving experiences~\cite{azees2016comprehensive}. However, the increased connectivity and data exchange among vehicles, infrastructure, and other devices have raised concerns about privacy preservation. To address the challenge, various techniques have been proposed with some focus on cryptographic schemes and anonymous authentication mechanisms~\cite{ni2016privacy}, while others explore the potential of machine learning and blockchain for privacy preservation in IoV~\cite{kumar2021privacy}.

\section{Methodology}

We first provide an overview of our approach. 
We the describe the reinforcement learning-based intersectional traffic coordination method~\cite{wang2023learning} as the preliminaries of our framework. 
Next, we introduce our privacy-protecting crowdsensing mechanism and present the graph propagation-based traffic forecasting algorithm.  
Lastly, we detail our route selecting method of RVs. 

\subsection{Framework Overview}
\begin{figure}[ht]
    \centering
    \includegraphics[width=\linewidth]{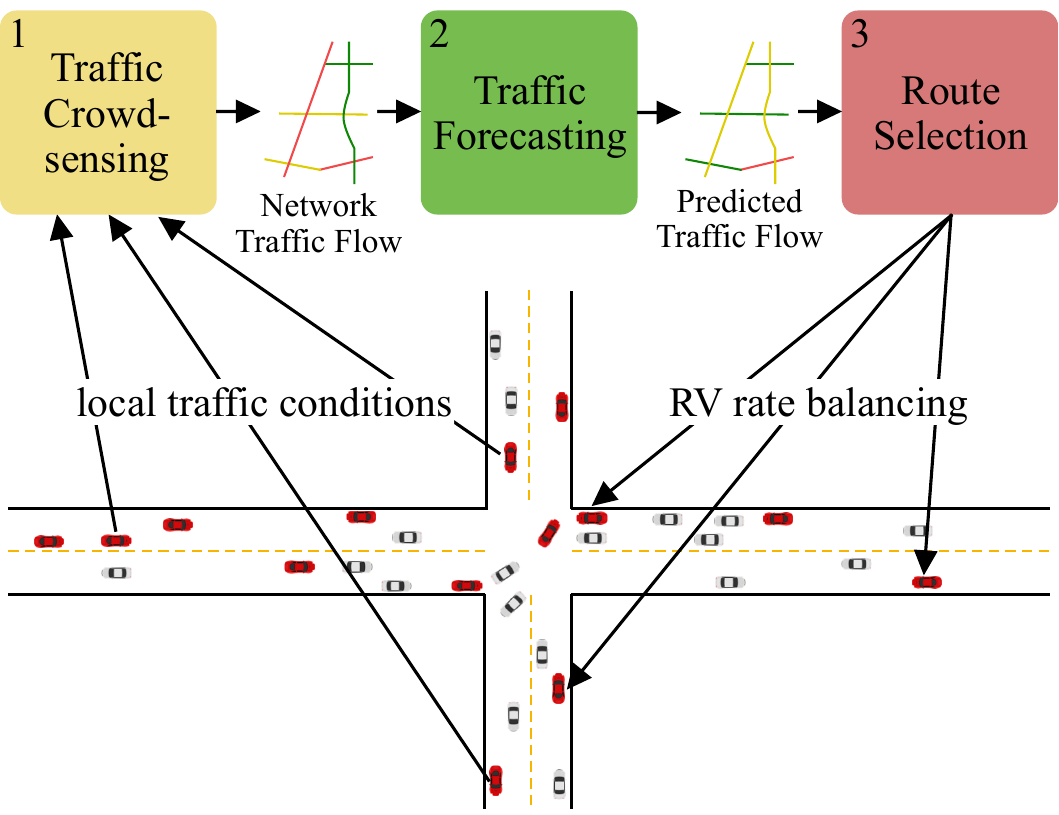}
    \caption{\small{An overview of our framework. (1) The traffic crowdsensing system collects data through RVs and constructs traffic flow of the entire road network. (2) The traffic forecasting method predicts potential traffic conditions. (3) The predicted traffic is adopted by the route selection mechanism to assign RVs to balance RV rates across the road network. When entering an intersection, RVs will employ an RL-based control policy to coordinate traffic and enhance overall traffic efficiency.}} 
    \label{fig:pipeline}
    \vspace{-1em} 
\end{figure}

Our framework consists of three modules:  
1) a privacy-protecting crowdsensing method, 
2) a graph propagation-based traffic forecasting method, 
and 3) a privacy-preserving route selection mechanism. Fig.~\ref{fig:pipeline} illustrates the pipeline of our system.
First, the crowdsensing method is adopted by each RV to collect local traffic conditions, which are then used to construct the traffic flow of the entire road network. 
Next, the constructed traffic flow is fed into the graph propagation-based traffic forecasting method~\cite{li2017diffusion} to produce a 100-step forecast for detecting the RV shortage road segments. 
In the following, the route selection mechanism balances RV rates on road segments via designated RVs, which traverse road segments experiencing RV shortage, to their destinations. 
Finally, RVs employ their reinforcement learning-based control policy to coordinate traffic at unsignalized intersections~\cite{wang2023learning}.

\subsection{Intersectional Traffic Coordination}
\begin{figure}[ht]
    \centering
    \includegraphics[width=.8\linewidth]{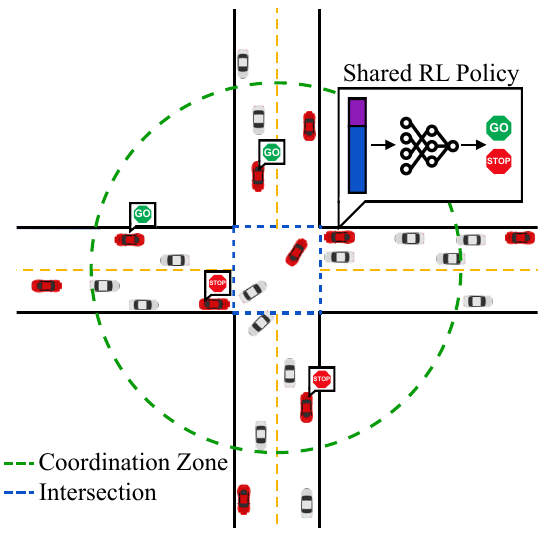}
    \caption{\small{The RL-based intersectional traffic coordination method~\cite{wang2023learning}. The intersection's traffic conditions are encoded, providing input to the RL policy used by the RV to determine Go/Stop upon reaching the intersection.}}
    \label{fig:prelim}
\end{figure}

We formulate intersectional traffic coordination as partially observable Markov decision process (POMDP), which can be solved by an reinforcement learning (RL) policy $\pi_\theta$ represented by a deep neural network~\cite{wang2023learning}.

The action space of $\pi_\theta$ is discrete and contains the decisions of whether an RV $i$ shall or shall not pass the entrance line of an intersection:
\begin{equation}
     a_t^i \in A = \{ \text{Stop}, \text{Go} \}.
\end{equation}

\noindent The observation space contains the status of the ego RV $d_t^i$, which denotes the distance between the ego RV $i$ and the entrance line of the intersection at time $t$, traffic condition inside the intersection $m_t^j$, which is an occupancy map for each moving direction $j$ at time $t$, and traffic condition outside the intersection, which includes the queue length $l_t^j$ and the average waiting time $w_t^j$ of each queue $j$ at time $t$: 
\begin{equation}
    o_t^i = \oplus_j^J \langle {l}^j_t, {w}^j_t \rangle \oplus_j^J \langle {m}_t^j \rangle \oplus \langle d_t^i \rangle,
\end{equation}
where $\oplus$ is the concatenation operator and $J = 8$ is the number of traffic moving directions at the intersection shown in Fig.~\ref{fig:prelim}.

\noindent The reward function is defined as 
\begin{equation}
    r(s^t, a^t, s^{t+1}) = \lambda_L r_L + p_c, 
\end{equation}
where $r_L$ is the local reward and $p_c$ is the conflict punishment. The local reward $r_L$ is the following:  
\begin{equation} 
\begin{cases}
  -{w}^{t+1,j},  & \ \text{if } {a}^t = \text{Stop};\\
  {w}^{t+1,j}, & \ \text{otherwise}.\\
\end{cases}
\label{eq:ego_reward}
\end{equation}
${w}^{t+1,j}$ is the average waiting time of all vehicles in the $j$th direction, which is normalized to $[0, 1]$.
If the RV opts to Stop, it incurs a negative waiting time $-{w}^{t+1,j}$; 
otherwise, it is positive ${w}^{t+1,j}$. 
$p_c$ is the penalty term where an RV's movement is in conflict with other vehicles at the intersection.

Once entering the coordination zone (see Fig.~\ref{fig:prelim}), RVs are controlled by the RL-based intersectional traffic coordination method. The traffic condition is encoded into a fixed-length representation, which serves as the observation space. This representation is then used as input to the RL policy, which determines whether the RV should Go or Stop. 
Lastly, a coordination mechanism that ensures conflict-free movements within the intersection is also implemented~\cite{wang2023learning}. 
For details on the intersectional traffic control and coordination mechanism, we refer interested readers to Wang et al.~\cite{wang2023learning}.

\subsection{Privacy-protecting Crowdsensing}
\label{sec:traffic_perception}
To obtain real-time traffic information, we adopt the sensing capability of RVs. However, directly sharing local observations, including surrounding vehicle types and trajectories among all RVs, subjects to privacy breach. 
To protect the privacy of perceived vehicles, RVs will only share the observed RV penetration rates. Other sensing data will be processed and stored locally. 
By doing this, the privacy of other vehicles gets protected during the data collection process.
In our experiments, vehicles within the $30~m$ range of each RV are observed. 
Each RV records the total number of surrounding vehicles along with their type (HV or RV). 
The computed RV penetration rate is sent to the central coordinator and shared among all RVs.

Upon receiving the traffic information as a result of crowdsensing, the central coordinator will compute the RV penetration rate of an entire road segment ($P_{e}$) and estimate the total number of vehicles on the road segment ($N_{e}$) as follows:
\begin{equation}
    P_{e} = \frac{\sum_{v \in RV_e} (P_v)}{\left|RV_e\right|}, \  N_{e} = \frac{{\left|RV_e\right|}}{P_{e}},
\end{equation}
where all $v \in RV_e$ makes the set of RVs on $e$, $P_v$ denotes the observed RV penetration rate reported by $v$, and $\left|RV_e\right|$ indicates the total number of RVs in the set of $RV_e$.
By sending only decentralized data, such as the RV penetration rate, to the central coordinator, the privacy of the local traffic information is preserved.

\subsection{Graph Propagation-Based Traffic Flow Forecasting}
\label{sec:forecasting}



\begin{figure}[ht]
    \centering
    \includegraphics[width=\linewidth]{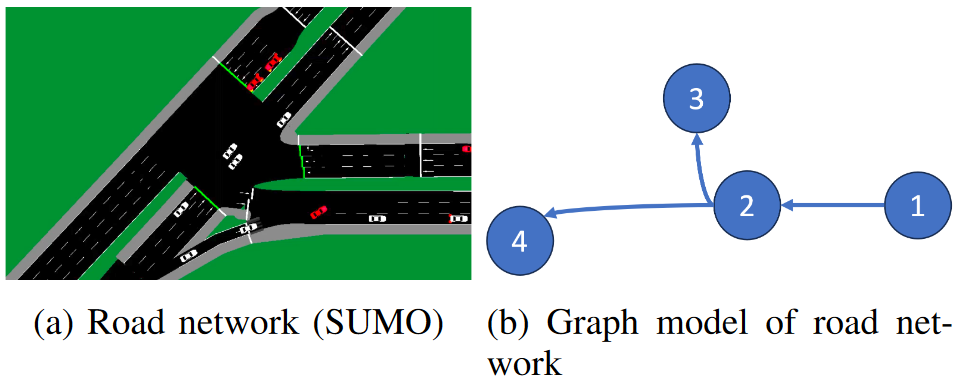}
    \caption{\small{Graph modeling a road network for traffic forecasting. (a) The original road network simulated in SUMO~\cite{behrisch2011sumo}. (b) The graph model of the road network. }}
    \label{fig:modeling}
\end{figure}

\subsubsection{Problem Formulation}
The goal of traffic forecasting is to predict not only traffic flow but also the RV penetration rate based on the observed traffic flow on a given road network. 
We represent the road network as a directed graph: 
\begin{equation}
    \mathcal{G} = (E, C, W),
\end{equation}

\noindent where ${E}$ is the set of road segments represented by the nodes of the graph, ${C}$ is a set of connections between each road segment represented by the edges of the graph, and ${W}$ is the adjacency matrix representing the connectivity between each pair of nodes. The observed traffic flow is denoted as ${X} \in {R}^{|E|\times F}$, where $|E|$ is the number of nodes in the graph, and $F$ represents features of mixed traffic, including the RV penetration rate of each road segment. 
The forecasting goal is to learn a function ${f}$ that predicts future traffic flow ${X}_{t+1}$ given the current traffic flow ${X}_t$ and the graph ${G}$: 
\begin{equation}
    {f}({X}_t;{G}) = {X}_{t+1}.
\end{equation}

\subsubsection{Graph Propagation Algorithm}
By representing the road network as a directed graph, each road segment has at least one predecessor.
An example is shown in Fig.~\ref{fig:modeling}. 
Therefore, we can define graph propagation as:
\begin{equation}
    X_{t+1}[e] = \sum_{p \in e.\text{pred}} (\alpha_p \cdot X_t[p]) + c.
\end{equation}
Here $X_{t+1}[e]$ is the traffic flow of the road segment $e$ at time $t+1$. Its value after graph propagation can be computed as a linear combination of $X_t[p]$ and the traffic flow of $e$'s predecessor $\{e.\text{pred}\}$. 
The coefficient and the intercept term of the linear combination are represented by $\alpha$ and $c$, respectively. 
To estimate the parameters of the linear model, we use data collected from our simulation. 
We minimize the residual sum of squares between the ground truth (simulation data) and the prediction of the linear approximation. 
Multi-step forecasting is achieved by consecutively running the forecasting model.

\subsection{Privacy-preserving Route Selection}
In order to balance the RV penetration rates across the road network so that RVs can be leveraged to smooth traffic, we need to route RVs from the road segments that have high RV rate to the road segments that have low RV rate. 
We propose a novel algorithm to achieve this balancing task by incorporating RV rates at all intersections for optimal route selection of RVs. 
We aim to balance RV rates in all moving directions of an intersection.
In addition, we ensure the privacy of the route selection process. 
Sensitive information such as the trajectory of an RV is kept private. 

\subsubsection{Overview} 
\begin{figure}[ht]
    \centering
    \includegraphics[width=\linewidth]{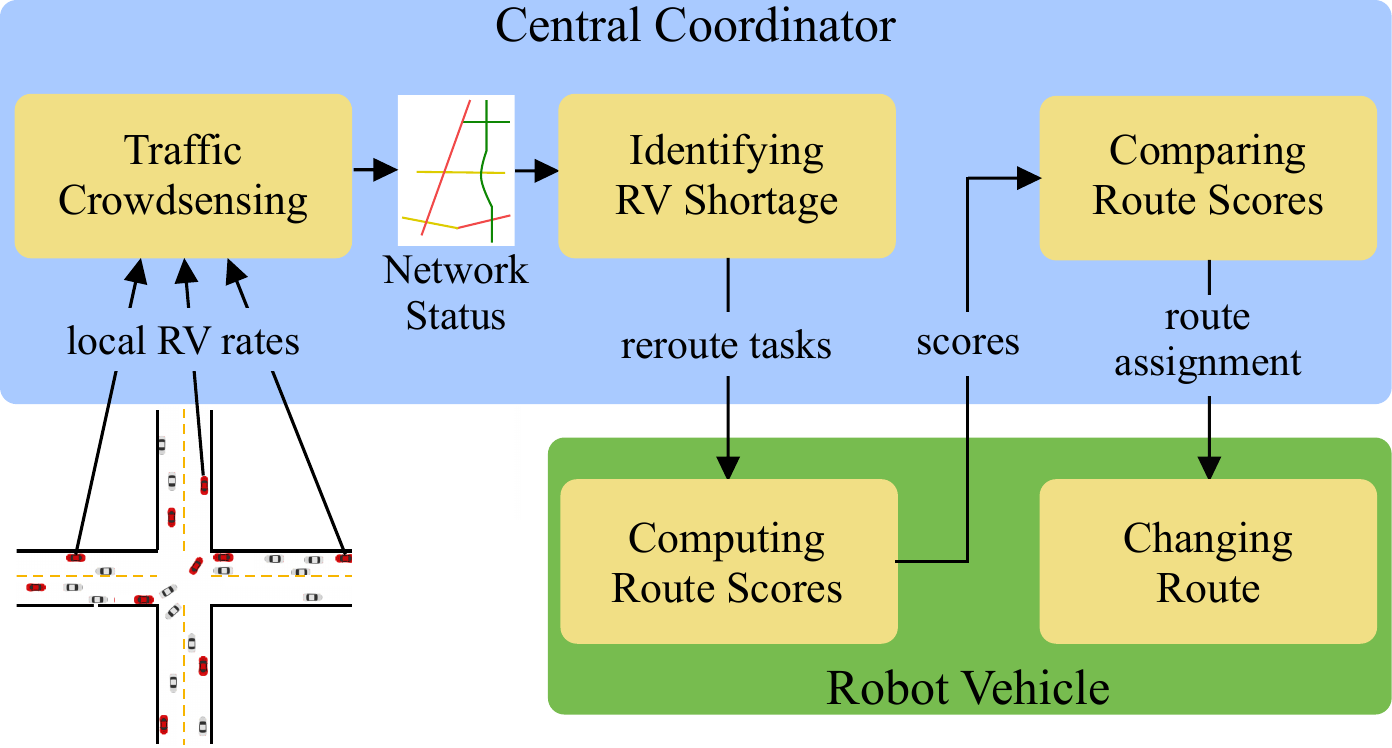}
    \caption{\small{The pipeline of the privacy-preserving route selection mechanism. The central coordinator first collects local RV rates via crowdsensing, then identifies road segments experiencing RV shortage and informs RVs. When RVs receive a reroute task, they will a score associated with each new route, which is sent back to the central coordinator for comparison and route assignment. }} 
    \label{fig:route_pipeline} 
\end{figure}

Fig.~\ref{fig:route_pipeline} illustrates our privacy-preserving route selection mechanism. The upper part of the figure depicts the data processing undertaken by the central coordinator, while the lower part represents the execution process of individual RVs.
The arrows represent communications between RVs and the central coordinator. 
First, RVs report their observed local RV rates to the central coordinator via crowdsensing.
Then, the central coordinator constructs traffic of the entire road network and identifies road segments that experience RV shortage. 
Afterwards, the central coordinator calculates the required number of vehicles for mitigating the shortage and re-route RVs for rate balancing. 
When an RV receives the re-routing task from the central coordinator, it will calculate a score of each new route and send it back to the central coordinator. 
The scores will be compared by the central coordinator and a new route will be designated for the RV.

Our method is designed to be privacy-preserving. 
The crowdsensing process ensures the local observation made by the RV is secured, and
the route selection process ensures the trajectory (including the destination) of an RV is private. 
No third party has the access to private information throughout the two processes.

\begin{figure}[ht]
    \centering
    \includegraphics[width=\linewidth]{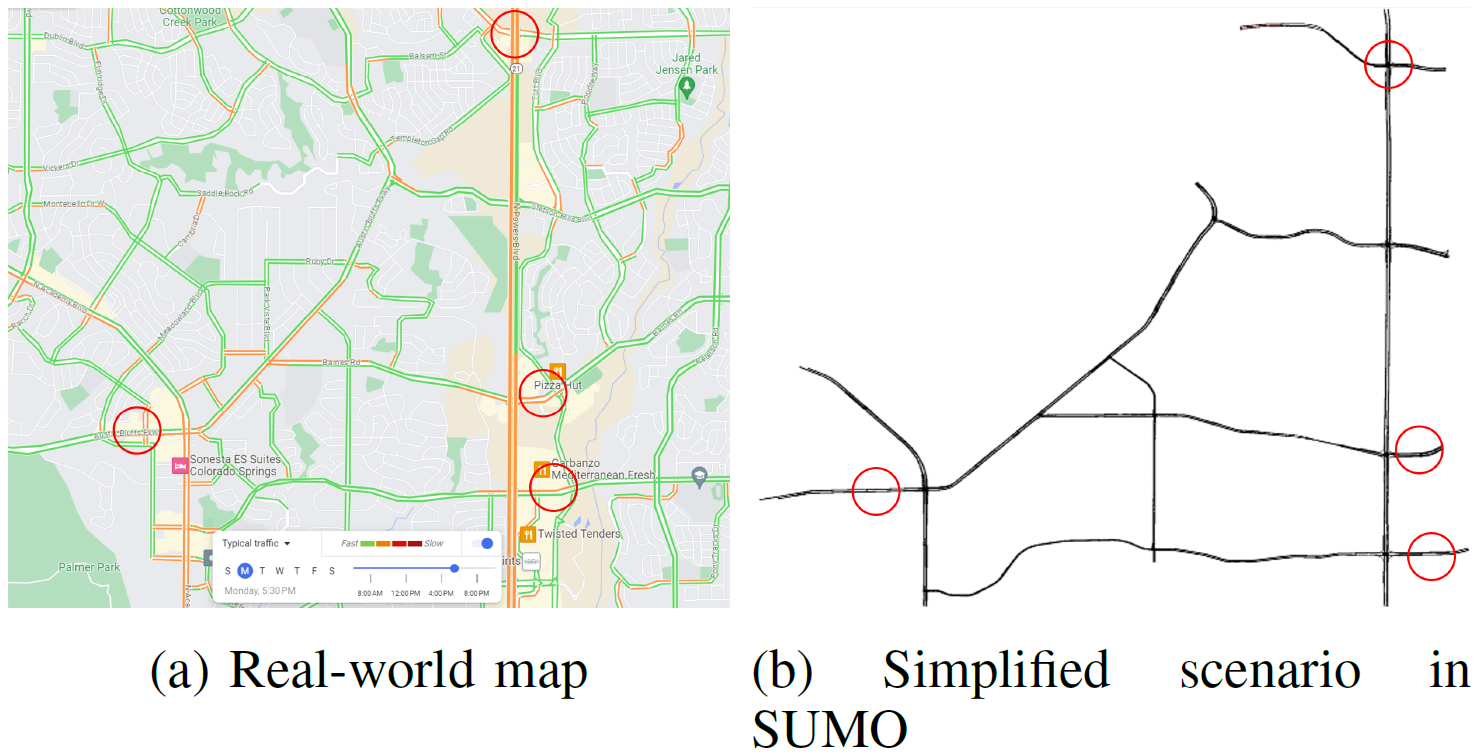}
    \caption{\small{The test area in the city of Colorado Springs, CO, USA. (a) Typical traffic at 5:30PM acoording to Google Map is shown. 
    (b) The simplified map with all secondary roads removed. Locations circled in red are assigned to be the low-RV-rate spawn points. }} 
    \label{fig:map}
    \vspace{-2em} 
\end{figure}

\subsubsection{RV Shortage Detection} 

The central coordinator uses estimated traffic condition to identify road segments with RV shortage. 
A road segment $e$ is marked having the shortage issue if the RV rate is below a given threshold: $$P_e < P_{\text{target}} - \lambda,$$ 
\noindent where $P_e$ is the RV rate of road segment $e$, $P_{\text{target}}$ is the target RV rate we want to achieve on all road segments of the road network, and $\lambda$ is a hyperparameter representing the shortage threshold and set to 5\% empirically. 
Thus, the desired number of vehicles to be sent to the shortage road segment $e$ is 
\begin{equation}
    D_e = \lceil (P_{\text{target}} - P_e)*N_e \rceil,
\end{equation}
where $N_e$ is the number of vehicles on $e$. 
If the RV rate of the predecessor road segment $e_{\text{pre}}$, i.e., $P_{\text{pre}}$, is greater than $P_{\text{target}}$, a re-routing task containing $e$ will be sent to RVs on $e_{\text{pre}}$.  

\subsubsection{Route Planning} 

Once a re-routing task is received, the RV will plan the shortest route to its destination via the requested RV shortage road segment. 
A corresponding score is computed for the new route:
\begin{equation}
\begin{split}
    \mathcal{S}(\text{new\_route}) = & \ \textsc{Dis}(\text{shortage\_road}) \\
             &+ \textsc{Len}(\text{new\_route}) \\
            &- \textsc{Len}(\text{curr\_route}),
\end{split}
\end{equation}
where $\textsc{Dis}(\text{shortage\_road})$ is the distance from the RV's current location to the requested road segment.
$\textsc{Len}(\text{new\_route}) - \textsc{Len}(\text{curr\_route})$ is the difference in length between the new route and the current route of the RV.

One issue is that a road segment may have more than one successor (see Fig.~\ref{fig:modeling}b for an example). 
This can result in RVs on such road segments receiving multiple re-routing tasks. 
In this case, the RV will compare the scores of all tasks and keep the task that has the minimum score. 
For tasks that have scores greater than the minimum score, their scores will be set to $\infty$ to prevent them from being selected.
Afterwards, the scores will be sent to the central coordinator for re-routing task assignment. 
The entire process is described in Algorithm~\ref{alg:route_planning}.
\begin{algorithm}
\begin{algorithmic}[1]
\State \textbf{Input:} a set of reroute\_tasks

smallest\_ID, smallest\_score $\gets$ $-1$, $\infty$ 
\For{task $\in$ reroute\_tasks}
\State new\_route $\gets$ shortest route via task.shortage\_road 
\State response.score $\gets \mathcal{S}(\text{new\_route})$
\State response.ID, response.veh\_id $\gets$ task.id, task.veh\_id
\If{smallest\_score $\geq$ response.score}
\State smallest\_score $\gets$ response.score
\State smallest\_ID $\gets$ response.id
\EndIf
\EndFor

\For{response $\in$ responses}
\If{ response.ID $\neq$ smallest\_ID}
\State response.score $\gets$ $\infty$
\EndIf
\EndFor
\State \textbf{Return:} responses with scores (to be sent to the central coordinator) 

\end{algorithmic}
\caption{Route Planning of RVs}
\label{alg:route_planning} 
\end{algorithm}

\subsubsection{Route Selection}
Once the central coordinator receives responses from the vehicles on $e_{\text{pre}}$, it organizes the responses into responses\_set$_e$, selects the lowest score based on the desired vehicle number $D_e$, and sends the decisions back to the corresponding RV for re-routing. 
The process is summarized in Algorithm~\ref{alg:route_select}, where \textsc{SortByScore} denotes the function of sorting the responses\_set by the route scores in an ascending order, and \textsc{Send}($a$, $b$) presents the process of sending message $b$ to the vehicle $a$.
\begin{algorithm}
\begin{algorithmic}[1]
\State \textbf{Input}: responses\_set$_e$ for shortage road segment $e$ 
\State \textsc{SortByScore}(responses\_set$_e$)
\For {$n \in [1,2,3,...,D_e]$}
\State \textsc{Send}(responses\_set$_e$[$n$].veh$\_$id, `change\_route'),
\EndFor
\end{algorithmic}
\caption{Route Selection via Central Coordinator}
\label{alg:route_select} 
\end{algorithm}

\section{Experiments and Results}
We begin by introducing our large-scale traffic simulation used in evaluation. 
We then present the results of the proposed traffic forecasting algorithm, followed by the evaluation results of the route selection mechanism. 
Lastly, we demonstrate large-scale mixed traffic controlled and coordinated by our approach.

\subsection{Large-scale Traffic Simulation}
\label{sec:traffic_sim}
To simulate large-scale mixed traffic, we use the microscopic traffic simulator SUMO~\cite{behrisch2011sumo}. The study area is the city of Colorado Springs, CO, USA, shown in Fig.~\ref{fig:map}a. 
After removing all secondary roads (see Fig.~\ref{fig:map}b), we select vehicle starting and ending edges at the border of the test area. 
Using SUMO's jtrrouter\footnote{https://sumo.dlr.de/docs/jtrrouter.html}, we generate the shortest routes for each pair of starting and ending edges, and automatically generate traffic flows in-between. 
It is worth noting that since the traffic flow is not reconstructed using real-world traffic data, the mixed traffic pattern and performance are not comparable to the results reported in the previous mixed traffic control study~\cite{wang2023learning}. 
In our test scenario, all intersections are either three-way or four-way. 
In addition, only left turn, right turn, and cross are allowed at the intersections. 
U-turn is prohibited for all vehicles. 

In order to simulate representative traffic patterns, we design three test scenarios based on the typical traffic conditions at three time periods of a day found in Google Map\footnote{https://www.google.com/maps}: 10AM (morning rush hour), 5:30PM (evening rush hour), and 10PM (evening hour). 
As shown in Fig.~\ref{fig:map}a, the traffic volumes on different road segments are varied. 
The road segments with large traffic volume during evening rush hours are circled (see Fig.~\ref{fig:map}b). 
Since the high volume is likely caused by off-work crowd, we assume a surge of HVs on the road network, causing dramatic drops of RV rates on these busy road segments, which naturally serve as low-RV-rate spawn points.

\subsection{Traffic Forecasting}
\subsubsection{Baselines and metrics}
We compare the effectiveness of our graph propagation-based traffic forecasting method with several commonly used traffic forecasting methods:
\begin{itemize}
    \item CONST: constant prediction, which assumes that traffic flow will remain stable and constant throughout the forecasting horizon.
    \item VAR: Vector Auto-Regression~\cite{hamilton2020time} that is implemented in the python package statsmodel\footnote{https://www.statsmodels.org/stable/index.html}. 
    \item DCRNN~\cite{li2017diffusion}: The state-of-the-art diffusion convolutional recurrent structure that leverages directed graph for traffic flow forecasting.
\end{itemize}
We use the following three metrics to evaluate the accuracy of our forecasting results:
\begin{itemize}
    \item Mean Absolute Error (MAE)
        \begin{equation}
            MAE = \frac{\sum_{i=1}^n \left| y_i -\hat{y}_i\right|}{n},
        \end{equation}
    \item Root Mean Squared Error (RMSE)
        \begin{equation}
        RMSE = \sqrt{\frac{\sum_{i=1}^n ( y_i - \hat{y}_i )^2}{n}},
        \end{equation}
    \item Mean Absolute Percentage Error (MAPE)  
            \begin{equation}
        MAPE = \frac{1}{n}\sum_{i=1}^n \left|\frac{ y_i-\hat{y}_i }{y_i}\right|.
        \end{equation}
\end{itemize}
The variables $y_i$ and $\hat{y}_i$ correspond to the actual and predicted values, respectively, and $n$ represents the length of the data.

\subsubsection{Datasets and results}
We conduct experiments over 20 simulation runs with each run lasting 2000 steps. 
We use simulation data to train and test our forecasting model: 70\% of the data for training, 20\% for evaluation, and 10\% for validation.
We aggregate the RV penetration rate on each road segment into time-series windows.

Table~\ref{tab:forecasting_comparison} compares the performance of the traffic forecasting methods using 10, 50, and 100 steps as the prediction horizon.
All predictions use 1-step history as input. 
Long-term forecasting is achieved by running the prediction models consecutively.
The experiments show that 
our method better captures network-wide evolution of traffic flow by 
outperforming the three baseline methods in both short-horizon and long-horizon traffic forecasting. 
The accurate predictions form a foundation for all subsequent processes of our framework.

\begin{table}[]
\caption{\small{Performance comparison of traffic forecasting methods. } }
\centering
\begin{tabular}{|c|c|c|c|c|c|}
\hline 
Horizon                    & Metrics & CONST & VAR~\cite{hamilton2020time}   & DCRNN~\cite{li2017diffusion}
          & Ours \\ \hline\hline
\multirow{3}{*}{10 step}  & MAE     & 0.898 & 0.099 & 0.069 &\textbf{0.068}\\ \cline{2-6} 
                           & RMSE    & 1.364 & 0.201 & 0.187 & \textbf{0.117}\\ \cline{2-6} 
                           & MAPE    & 1.097 & 0.123 & 0.093 & \textbf{0.092}\\ \hline
\multirow{3}{*}{50 steps}  & MAE     & 1.102 & 0.125 & 0.105 & \textbf{0.101}\\ \cline{2-6} 
                           & RMSE    & 1.672 & 0.252 & 0.237 & \textbf{0.138}\\ \cline{2-6} 
                           & MAPE    & 1.341 & 0.174 & 0.149 &\textbf{0.138}\\ \hline
\multirow{3}{*}{100 steps} & MAE     & 1.505 & 0.145 & 0.119 &\textbf{0.109}\\ \cline{2-6} 
                           & RMSE    & 2.283 & 0.273 & 0.251 &\textbf{0.144}\\ \cline{2-6} 
                           & MAPE    & 1.824 & 0.195 & 0.167 &\textbf{0.150} \\ \hline
\end{tabular}
\label{tab:forecasting_comparison}
\vspace{-1em}
\end{table}


\begin{figure*}[ht]
    \centering
    \includegraphics[width=\linewidth]{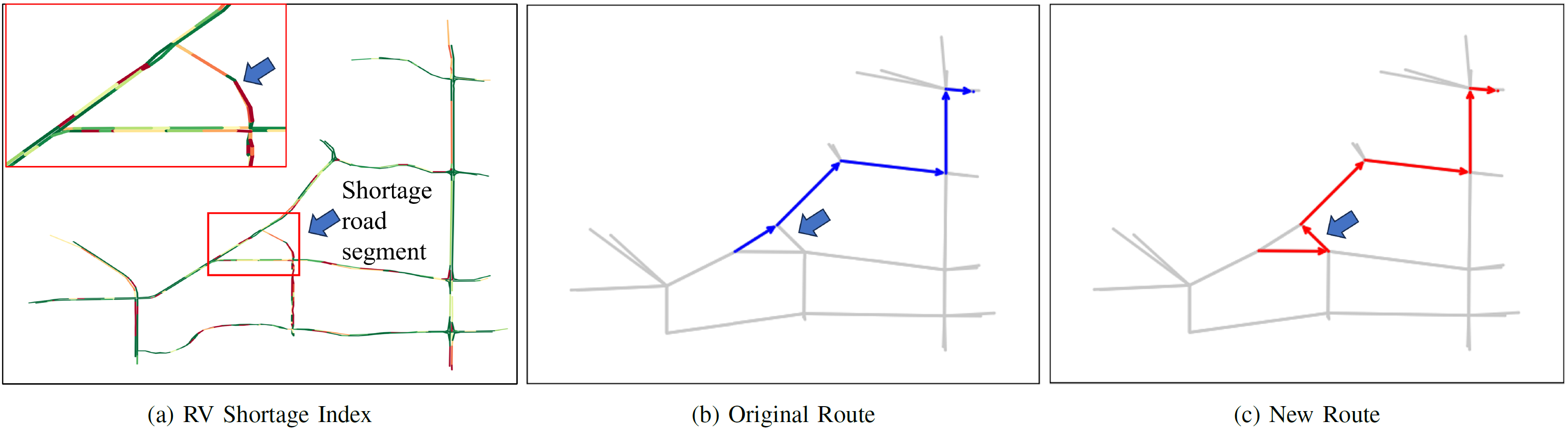}
    \caption{\small{An example of re-routing an RV. (a) The RV shortage index of each road segment is shown and re-routing tasks of RVs are created. (b) The original route of a designated RV. (c) The proposed new route of the designated RV for mitigating the RV shortage issue of the marked road segment.}} 
    \label{fig:case1}
\end{figure*}


\begin{figure*}[ht]
    \centering
    \includegraphics[width=\linewidth]{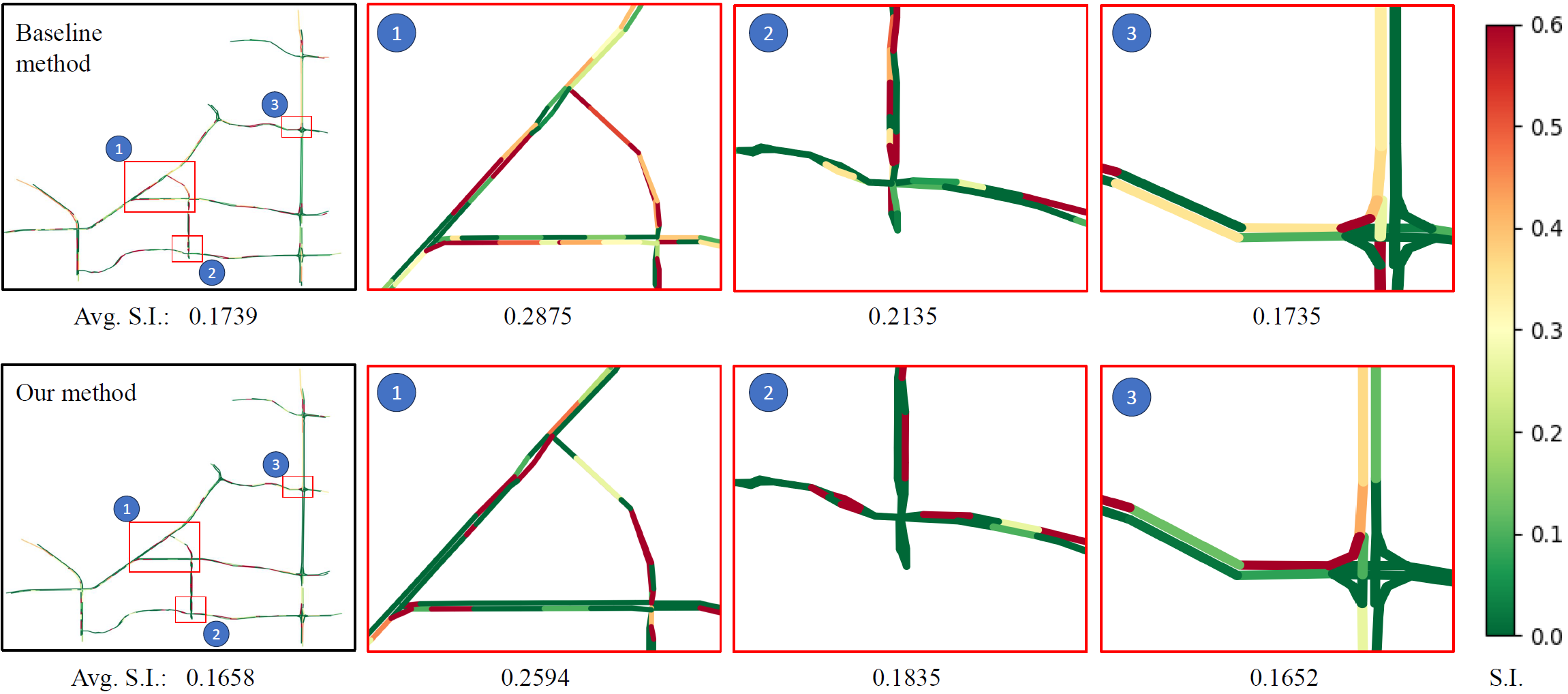}
    \caption{\small{The RV shortage issue of the road network along with zoomed-in views of three critical areas. Our method (bottom figures) shows apparent advantages of mitigating the RV shortage issue (measured in average RV shortage index, i.e., Avg. S.I.)  over the baseline method (top figures).}} 
    \label{fig:edge_shortage_illustration}
    \vspace{-1em}
\end{figure*}


\begin{figure}[ht]
    \centering
    \includegraphics[width=\linewidth]{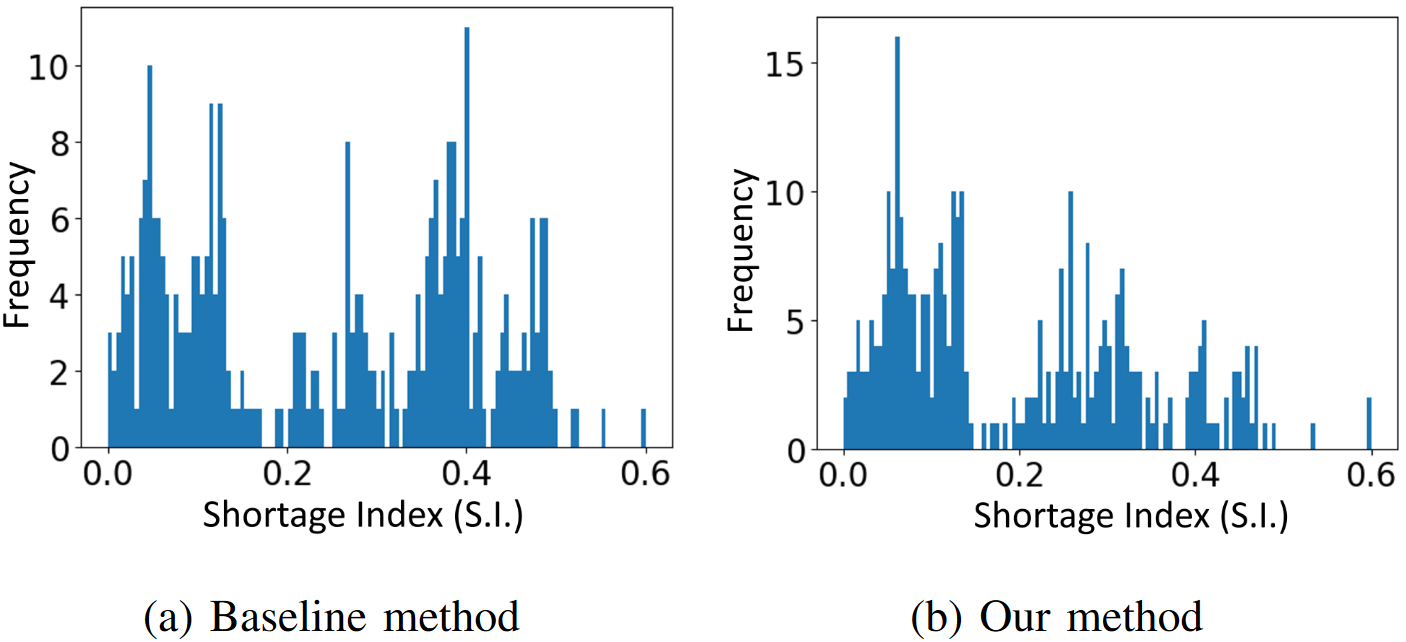}
    \caption{\small{The spatial distribution of shortage indices over the road segments of the test road network. The x-axis is the shortage index and the y-axis shows the frequency. Our method effectively mitigates the RV shortage issue of the road network indicated by more indices concentrating in lower-scoring areas. }} 
    \label{fig:edge_hist}
    \vspace{-1em}
\end{figure}

\subsection{Route Selection}
\label{sec:route_selection_result}
We evaluate our route selection mechanism by comparing it to several other route planning algorithms. 
We use the shortest route planner as the baseline. 
During evaluation, we employ either our method or the shortest route planner. 
The RV penetration rates of all road segments are recorded, and the RV shortage index (Shortage Index or S.I.) used in evaluation is defined as follows:
\begin{equation}
    \text{shortage\_index}(e) = 
        \begin{cases}
  0  & \ \text{if }  P_{e} > P_{\text{target}}, \\
  P_{\text{target}}-P_{e} & \ \text{otherwise,}
\end{cases}
\end{equation}
where shortage\_index$(e)$ denotes the RV shortage index of road segment $e$, $P_{\text{target}}$ and $P_{e}$ are the predefined target RV penetration rate and the recorded RV penetration rate on $e$, respectively.

In our simulation, spawned vehicles will be randomly designated as either HV or RV based on a predefined RV penetration rate. 
We evaluate three target RV rates $P_{\text{target}} = \{50\%, 60\%, 80\%\}$. 
The corresponding spawn RV rates are set higher as $\{55\%, 65\%, 85\%\}$, since it is necessary to have a surplus of RVs for reallocation; otherwise, at least one road segment in the network will suffer from RV shortage.   
During evaluation, we set the RV rate to $20\%$ at several designated spawn points to mimic real-world traffic conditions as discussed in Sec.~\ref{sec:traffic_sim}.

Fig.~\ref{fig:case1} gives an example of re-routing to mitigate the RV shortage issue.  Fig.~\ref{fig:case1}a shows the RV shortage index for each road segment, with an arrow indicating the re-routing task of an RV. 
Fig.~\ref{fig:case1}b and Fig.~\ref{fig:case1}c illustrate the original and proposed routes of the designated RV, respectively. 
Shown in Fig.~\ref{fig:case1}a, the designated RV chooses the proposed new route via the road segment experiencing the RV shortage issue

The results illustrating the shortage indices of the road network after 900 steps are shown in Fig.~\ref{fig:edge_shortage_illustration}: the top figures display, using the baseline method, the severity of shortages across the entire road network as well as the zoomed-in views of three critical areas. 
The average RV shortage index (Avg. S.I.) of the three critical areas are also shown.  
In contrast, the bottom figures present the results of using our method. 
It is apparent that our method effectively mitigates the RV shortage issue across the road network.

Fig.~\ref{fig:edge_hist}a shows the shortage indices of using the baseline method are almost uniformly distributed from 0.0 to 0.5. In contrast, when our method is adopted, the shortage indices concentrate more in low-scoring areas, e.g., (0.0, 0.1) and (0.1,0.2), as shown in Fig.~\ref{fig:edge_hist}b. These results provide further evidence that our method effectively mitigates the RV shortage issue across the entire road network.

\begin{figure*}[htbp]
    \centering
    \includegraphics[width=\linewidth]{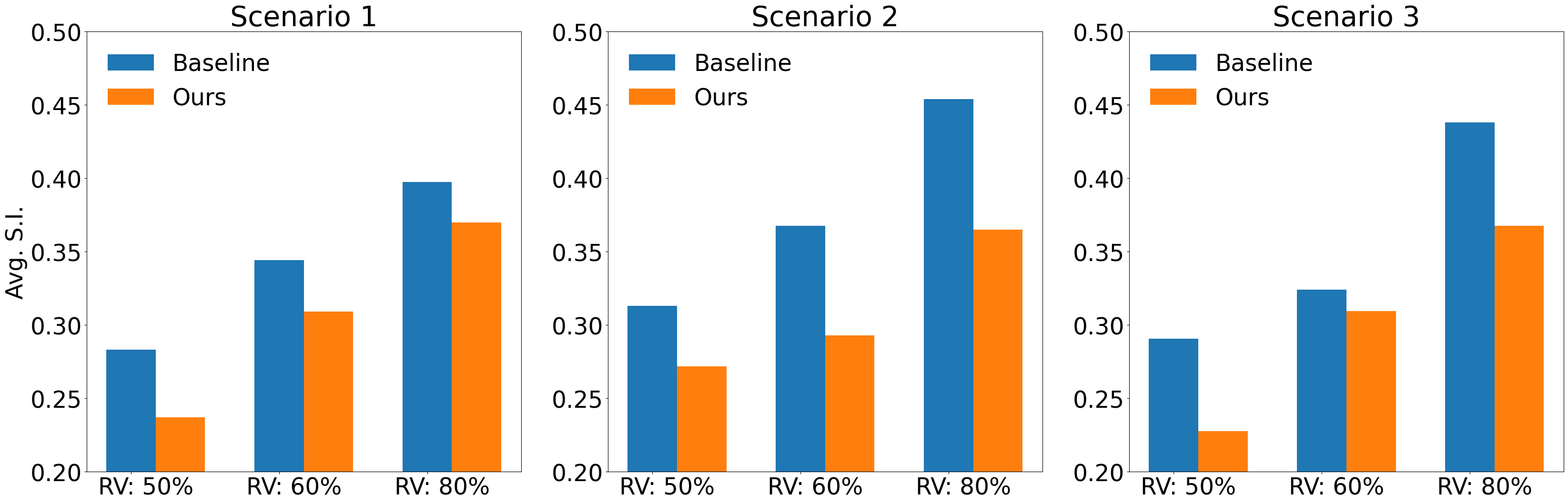}
    \caption{\small{Evaluation of our route selection mechanism. We report the average shortage index as a result of using our method vs. using the baseline method in three test scenarios. The results suggest that our method can mitigate the RV shortage issue and re-balance the RV rate across the entire road network. } } 
    \label{fig:shortplot}
    \vspace{-.5em}
\end{figure*}

Fig.~\ref{fig:shortplot} presents quantitative results of our route selection mechanism. 
The average RV shortage indices over all road segments are reduced as the RV penetration rate increases. 
A notable example is when the RV penetration rate is $50\%$, the average shortage index is reduced by $69.4\%$ in Scenario 3 via our method. 
As the RV shortage issue gets alleviated, more RVs can be leveraged to smooth traffic conditions of a road segment.

\begin{figure*}[htbp]
    \centering
    \includegraphics[width=\linewidth]{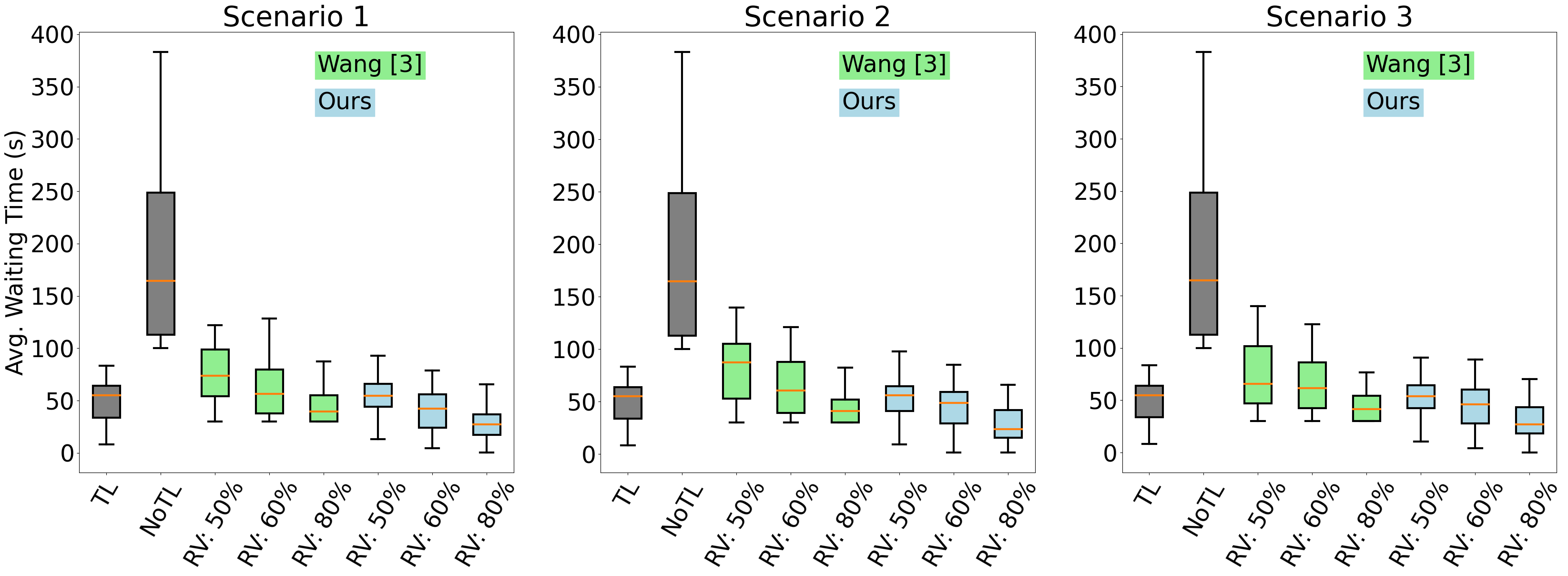}
    \caption{\small{Comparison of our method to other baseline methods in mixed traffic control. The x-axis denotes traffic coordination methods and the y-axis is the average waiting time of all vehicles in the test area. The results show that our method is the most effective among the evaluated methods.} }
    \label{fig:eval}
    \vspace{-1.5em}
\end{figure*}

\subsection{Large-scale Mixed Traffic Coordination}

We present the results of deploying our method in the entire test area by comparing it with three baseline methods:
\begin{itemize}
    \item NoTL: All traffic signals are off and no RV is present.
    \item TL: Traffic is coordinated using traffic signals.
    \item Wang~\cite{wang2023learning}: The state-of-the-art mixed traffic coordination method for single intersections.
\end{itemize}

We run each experiment for 1000 steps with the same RV penetration rate set for our method and Wang~\cite{wang2023learning}. 
In Fig.~\ref{fig:eval}, we present the average waiting time of all vehicles across the entire network.  
Our method not only outperforms Wang~\cite{wang2023learning} at the same RV penetration rates but also outperforms the traffic light baseline starting at 50\% RV rate. For instance, the average waiting time is reduced by $27\%$ when the RV penetration rate is $50\%$ in Scenario 3. 
In comparison, Wang~\cite{wang2023learning} starts to outperform the traffic light baseline when the RV penetration rate is $60\%$ or above.

\section{Conclusion and Future Work}
We propose a framework for large-scale mixed traffic control and coordination. 
Our framework consists of three novel methods: a privacy-protecting crowdsensing method for collecting local traffic conditions, 
a graph propagation-based traffic flow forecasting method for predicting RV penetration rates of the road network, 
and a privacy-preserving route selection mechanism for mitigating network-wide RV shortage issues.
Our framework has been evaluated using a real-world network. 
Extensive experiments show that our approach outperforms other baseline methods over all sub-tasks and substantially improves the overall efficiency of mixed traffic control and coordination.  

There are many future directions we can pursue.
First of all, federal learning can be adopted by our framework to form a privacy-preserving and context-aware routing algorithm. Second, city-scale traffic information can be combined with previous city-scale traffic reconstruction and optimization techniques~\cite{Wilkie2015Virtual,Li2017CityFlowRecon} to scale up our test scenarios. 
Finally, the integration of additional smart city technologies and devices for sensing and communication can further enhance the efficiency of mixed traffic control and coordination.

\section*{Acknowledgments}
Dawei Wang and Jia Pan are supported by the Innovation and Technology Commission of the HKSAR Government under the InnoHK initiative. Jia Pan is also supported by ITF GHP/126/21GD and HKU's CRF seed grant. 
Weizi Li is supported by NSF IIS-2153426. He would like to thank NVIDIA and the University of Tennessee, Knoxville for their support.
The authors would also like to thank Michael Villarreal for creating some of the figures.

\clearpage
\ifCLASSOPTIONcaptionsoff
  \newpage
\fi



%
\bibliographystyle{IEEEtran}
\bibliography{reference}




\end{document}